\UseRawInputEncoding

\documentclass[10pt,journal,compsoc]{IEEEtran}
\ifCLASSOPTIONcompsoc
  \usepackage[nocompress]{cite}
\else
  \usepackage{cite}
\fi
\usepackage{ragged2e}
\usepackage{graphicx}
\usepackage{amsfonts}
\usepackage{amsmath}
\usepackage{amssymb}
\usepackage{float}
\usepackage{wrapfig}
\usepackage{algpseudocode}
\usepackage{algorithmicx}
\usepackage{algorithm}
\usepackage{booktabs}
\usepackage{verbatim}
\usepackage{multirow}
\usepackage{threeparttable}
\usepackage{amsmath}

\ifCLASSINFOpdf
\else
\fi
\begin{document}
%
\title{Class-Aware Universum Inspired Re-Balance Learning for Long-Tailed Recognition}
%
%
%
%

\author{Enhao~Zhang$^{\ast}$,
        Chuanxing~Geng$^{\ast}$ \thanks{*The first two authors contributed equally to this work},
        and~Songcan~Chen
\IEEEcompsocitemizethanks{\IEEEcompsocthanksitem  E. Zhang, C. Geng and S. Chen are with MIIT Key Laboratory of Pattern Analysis and Machine Intelligence, China
and College of Computer Science and Technology, Nanjing University of Aeronautics and Astronautics (NUAA), Nanjing 211106, China\protect\\
E-mail:\{zhangeh, gengchuanxing, s.chen\}@nuaa.edu.cn.}
}

\IEEEtitleabstractindextext{%
\begin{abstract}
\justifying Data augmentation for minority classes is an effective strategy for long-tailed recognition, thus developing a large number of methods. Although these methods all ensure the balance in sample quantity, the quality of the augmented samples is not always satisfactory for recognition, being prone to such problems as over-fitting, lack of diversity, semantic drift, etc. For these issues, we propose the Class-aware Universum Inspired Re-balance Learning(CaUIRL) for long-tailed recognition, which endows the Universum with class-aware ability to re-balance individual minority classes from both sample quantity and quality. In particular, we theoretically prove that the classifiers learned by CaUIRL are consistent with those learned under the balanced condition from a Bayesian perspective. In addition, we further develop a higher-order mixup approach, which can automatically generate class-aware Universum(CaU) data without resorting to any external data. Unlike the traditional Universum, such generated Universum additionally takes the domain similarity, class separability, and sample diversity into account. Extensive experiments on benchmark datasets demonstrate the surprising advantages of our method, especially the top1 accuracy in minority classes is improved by $1.9\%  \sim  6\%$ compared to the state-of-the-art method.
\end{abstract}

}

\maketitle

\IEEEdisplaynontitleabstractindextext

%
\IEEEpeerreviewmaketitle

\IEEEraisesectionheading{\section{Introduction}\label{sec:introduction}}

\begin{figure}
 \begin{center}
   \includegraphics[scale=0.57]{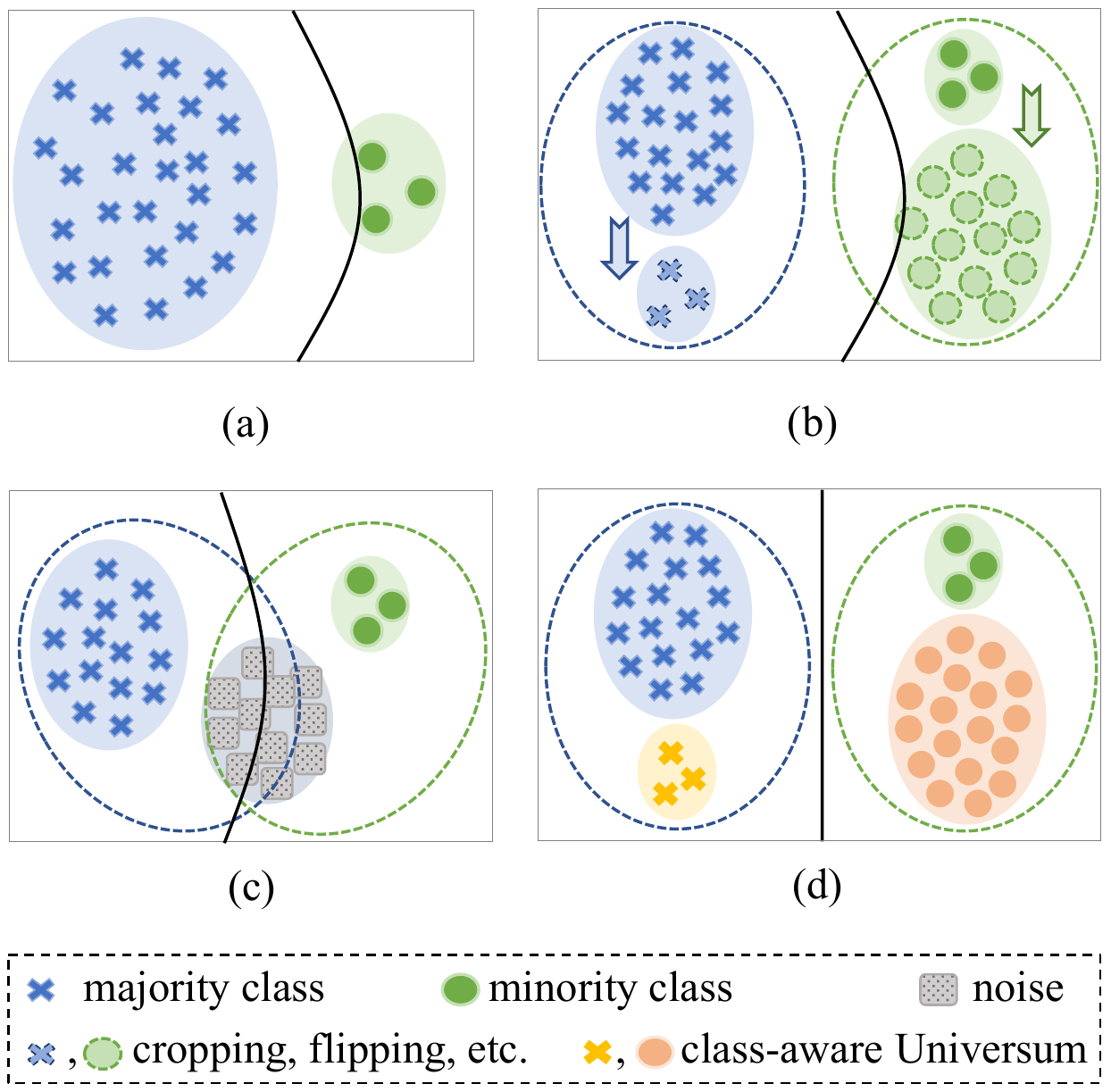}
   \caption{Binary classification long-tailed identification legend, where majority(blue fork) and minority(green circle) are relative, so the both of them need to be added by additional samples. \emph{(a)} The classification surface obtained directly employing long-tailed data cannot classify minority class. \emph{(b)} Traditional augmentation methods only achieve a balance in the number of samples, and the essence of the resulting classification surface is not improved. \emph{(c)} Because pure noise pictures are not discriminative, this method reduces the separability between different classes. \emph{(d)} Our method adds different classes of Universum (yellow and orange) to various classes of target data (blue and green), endowing the Universum class awareness to improve the performance.}

   \label{1}
 \end{center}
\end{figure}

%
%
%
%
\IEEEPARstart{D}{eep} neural networks(DNNs) have achieved a tremendous success in various computer vision tasks, which is inseparable from large-scale annotated datasets. The class distribution of these datasets is usually uniform, such as Cifar\cite{ref1}, ImageNet\cite{ref2}, COCO\cite{ref3}, etc. While the real-world data usually present a long-tailed distribution structure: a few classes are highly frequent, yet others are only rarely encoded\cite{ref4},\cite{ref5}. Such settings, for example, naturally arise in the data for common cerebrovascular diseases and rare hemophilia, which play a vital role in accurately detecting minority classes in comparable cases\cite{ref6}. However, learning the minority class could be quite challenging since the low-frequency classes can be easily overwhelmed by high-frequency ones\cite{ref7}. As shown in Fig.\ref{1}(a), the classifier (black curve) learned from imbalanced data easily gets biased by the majority class (blue fork), especially for DNNs-based models, easily resulting in poor classification performance. Therefore, long-tailed recognition has received a widespread attention recently.

Data augmentation is an effective strategy for long-tailed recognition, which aims to re-balance the data distribution by supplementing the number of minority class samples. Currently, a large number of data augmentation-based methods have been developed such as over-sampling \cite{ref8},\cite{ref24}, traditional augmentation(cropping, flipping, etc.)\cite{ref14},\cite{ref15}, \cite{ref16}, and introducing noise\cite{ref10},\cite{ref17},\cite{ref18}. Although these methods all ensure the balance in sample quantity (\emph{i.e.}, the number of samples ), the quality (like the sample diversity, etc.) of the augmented samples is not always satisfactory for recognition.

 Concretely, over-sampling only raises the sampling frequency of minority classes, which can not contribute to the diversity of their augmented samples. The traditional augmentation attempts to improve the diversity of augmented samples through cropping, flipping, etc. However, it is difficult to achieve this purpose due to the inherent limitation of minority classes, namely, the serious scarcity of samples. Though these methods improve the performance of long-tailed recognition somewhat, the classifiers learned may not be substantially different from those obtained directly from the original imbalance data, as illustrated in  Fig.\ref{1}(b).

To further increase the diversity of augmented samples, some researchers attempted to inject noise into the original samples \cite{ref10},\cite{ref17},\cite{ref18}. Despite of the boosting on model's performance, such strategy endures the following drawbacks: (1) Injecting a small bit of noise may fall into the dilemma described above. (2) Injecting too much noise may damage the semantic information of the original images. Recently, Zada et al.\cite{ref10} proposed OPeN(Oversampling with Pure Noise Images) by directly introducing pure noise images into different minority classes, and surprisingly the model performance was significantly increased, which seems to contradict the above drawback (2). But please note that they additionally introduced a DAR-BN(Distribution-Aware Routing Batch Normalization) technique, which can effectively alleviate the domain shift caused by the pure noise images introduced. Though this method has achieved the SOTA performance, it treats pure noise images identically, namely, adding pure noise images of the same distribution to different minority classes, which might further damage the separability of minority classes. As shown in Fig.\ref{1}(c), the added pure noise images (grey square) are indivisible, causing the generated classifier to be blind in categorizing it.

In this paper, we revisit the long-tailed recognition from a perspective of auxiliary contradictory  data, a.k.a Universum\cite{ref9}. It refers to the data not belonging to any category of the target tasks. Thus the added pure noise images used in OPeN can be actually viewed as the Universum of a special type because these pure noise images do not belong to any target classes. However, directly using Universum for long-tailed recognition is ineffective due to its class agnostics, that is, as a whole contradictory class for all given target classes, the existing Universum does not concern the class distribution, let the balance between target classes alone. To adapt the Universum strategy to long-tailed recognition, we innovatively propose a Class-aware Universum Inspired Re-balance Learning(CaUIRL), meaning that our Universum is introduced as the corresponding contradiction class to the individual target class. Unlike existing Universum strategies and OPeN, it is our CaUIRL that achieves the best of both worlds by not only rebalancing the individual minority classes but also promoting their separability. As in Fig.\ref{1}(d), we add certain category specific Universum according to the target class, endowing the Universum with the class-aware capability to aid and improve model performance. In particular, we theoretically demonstrate that the classifier learned by CaUIRL is consistent with the balanced scenario from a Bayesian viewpoint. Furthermore, unlike the previously stated approaches, CaUIRL neither restricts the training data to existing images and their augmentation nor introduces pure noise pictures, therefore bypassing their limitations.

In the Universum learning, how to choose the appropriate Universum is essential as well as an open question at present\cite{ref34}. To alleviate it, motivated by a popular mixup strategy, we novelly propose a Higher-order Mixup(HoMu) approach to automatically generate class-aware Universum(CaU) from given target data without resorting to any external data. Such CaU not only inherits the outstanding characteristics of traditional Universum but also alleviates the domain shift problem caused by the external data introduction. In addition, this strategy, \emph{i.e.}, automatically generating Universum, elevates Universum learning to a new level.

In summary, our key contributions and novelties are given as follows:
\begin{itemize}
\item To the best of our knowledge, we are the first to address the problem of long-tailed recognition from the Universum perspective and innovatively propose a framework for Class-aware Universum inspired re-balance learning.

\item For long-tailed scenarios, we additionally propose the \emph{corresponding class gap (C2G)} to measure the degree of domain shift between individual classes in different datasets, and according to C2G, the appropriate Universum can be efficiently selected.

\item From the perspective of Bayesian classification, we provide theoretical proof to guarantee that the classifier obtained using CaUIRL is consistent with the balanced scene.

\item We further propose a \emph{higher-order mixup} for automatically generating CaU without resorting to any external data, and HoMu itself has an independent interest as a kind of data augmentation.

\item Extensive experiments on a range of benchmark datasets demonstrate CaUIRL's superior performance, especially generalization on minority classes.
\end{itemize}



\section{Related Work}
In this section we briefly review the previous work closely related to our method, \emph{i.e.}, re-balance learning, Universum learning, and mixup.

\subsection{Re-balance Learning}
Long-tailed recognition seeks to achieve re-balancing by giving equal attention to the head (majority) classes and the tail (minority) classes\cite{ref19},\cite{ref20},\cite{ref21}. Specifically, the existing re-balancing strategies can be roughly divided into the following two facets.

\subsubsection{Loss Facet}
At the loss facet, the object is to assign more penalties to majority classes through the adjustment of loss values to mitigate the impact of imbalanced data. Common methods include re-weighting\cite{ref11},\cite{ref30},\cite{ref31}, margin loss\cite{ref6},\cite{ref12},\cite{ref33}, and confidence calibration\cite{ref13}.
The re-weighting idea is to re-weight classes proportionally by the inverse of their frequencies. Re-weighting classes directly is often insufficient, resulting in poor model performance. And in large-scale scenarios, re-weighting methods often make the optimization of deep models difficult\cite{ref28}. Marginal loss is used to modify the loss function to increase the margin for the minority classes. Confidence calibration handles different degrees of class overconfidence through label-aware smoothing, which can alleviate long-tailed recognition problems.

\begin{figure*}
 \begin{center}
   \includegraphics[scale=0.85]{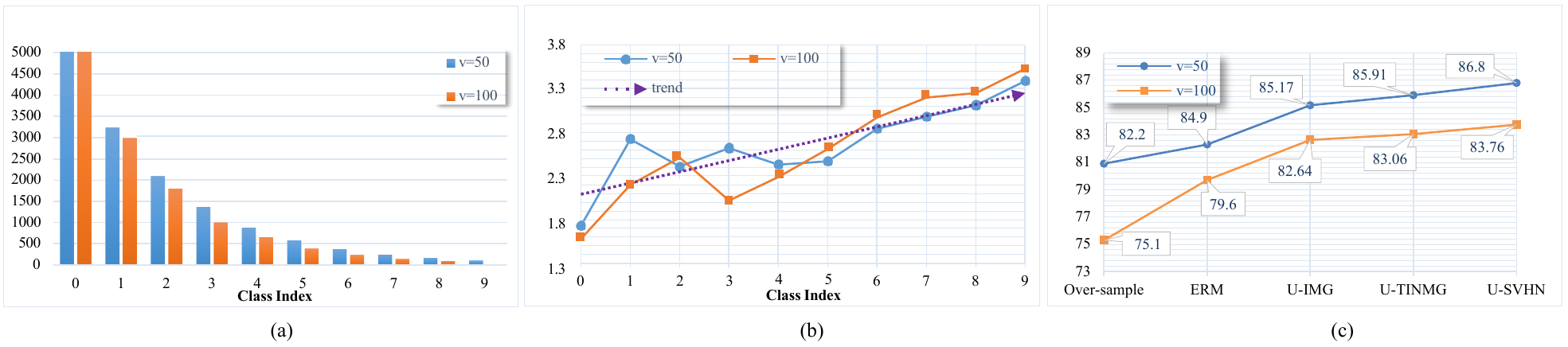}
   \caption{(a) is the distribution diagram of long-tailed data. (b) represents the relationship between the C2G and class index under different unbalance rates. (c) shows the classification performance of the five algorithms under different imbalance rates.}
   \label{2}
 \end{center}
\end{figure*}
\subsubsection{Sample Facet}
At the sample facet, it aims to re-balance the class prior distributions in a pre-processing procedure. Typical algorithms include under-sampling the head data\cite{ref7},\cite{ref22},\cite{ref23}, over-sampling the tail data\cite{ref8},\cite{ref24}, and adding/generating data\cite{ref10},\cite{ref17},\cite{ref26},\cite{ref27},\cite{ref41}. Generally speaking, these methods are simple and effective, but they also have some drawbacks involved. For example, both commonly used "over-sampling" and "under-sampling" strategies lead to over-fitting to tail classes because they cannot handle the lack of tail classes information\cite{ref41}. To enrich tail classes information, Kim et al.\cite{ref41} generate tail classes by translating samples from head classes. Wei et al.\cite{ref27} used open-set data to re-balance the class prior in an "\emph{Open-sampling}" manner. Zada et al.\cite{ref10} proposed the \emph{OPeN} framework, which employs pure noise images from the same Gaussian distribution as additional training samples for re-balancing. \emph{OpeN} is a method similar to our work, the difference lies in that our approach leverages auto-generated CaU without introducing external samples. Compared with pure noise images, CaU not only increases the diversity of minority classes, but also assists the target task with the help of class information, and the automatically generated CaU avoids the introduction of external data to alleviate domain shift. In particular, we theoretically demonstrate the effectiveness of our algorithm from a Bayesian perspective.

\subsection{Universum Learning}
Universum learning was first proposed by Vapnik\cite{ref34}, using Universum to maximize the classification margin by maximizing the number of observed contradictions, and presented Universum Support Vector Machine (USVM). Additionally, Chapelle et al.\cite{ref35} conducted a theoretical analysis of using Universum data in favor of SVM. As a new study scenario, Universum learning assists the target task by introducing appropriate data, but how to choose a suitable Universum is a challenge. The usual Universum data selection method is obtained by sampling external data with the same domain using techniques like $U_{rest}$\cite{ref49}, In-Between-Universum(IBU)\cite{ref50} and Information-Entropy-Based(IEB)\cite{ref37}. Naturally, capturing prior information on such exotic Universum unavoidably increases the computational cost. To alleviate this problem, Bai et al.\cite{ref51} used random averaging of samples to generate Universum data. Richhariya et al.\cite{ref38} used the information entropy of samples to generate Universum data. But the efforts mentioned above are only applicable to binary classification problems and are not universal. Dhar et al. \cite{ref36} extend Universum learning to multiclass learning. LeCun et al. \cite{ref36} extended the idea of Universum from support vector machines to deep learning, and proved that using Universum (sampling from unlabeled data) produces a certain gain effect on deep models.

In summary, the aforementioned Universum-based strategies do not take into account the target class distribution (\emph{i.e.}, class-agnostic), making such Universum incapable of coping with class imbalanced scenarios. To overcome this issue, we propose a class-aware Universum, combine the target class prior distribution with special processing of Universum, and design a completely different automatic generation method from the traditional heuristic construction of Universum.
\subsection{Mixup}
Mixup trains a neural network on convex combinations of pairs of examples and their labels\cite{ref16}. As a regularization technique, mixup has been applied in various tasks and exhibited better generalization and robustness. Based on the effectiveness of the mixup training method, the follow-up work is also exemplified by experiments \cite{ref13},\cite{ref40},\cite{ref52},\cite{ref53}. Driven by the excellent experimental performance of mixup, recent theoretical analysis based on mixup training has attracted attention. Zhang et al.\cite{ref54} demonstrate how mixup aids deep models to improve robustness and generalization through theoretical analysis. Harris et al.\cite{ref39} analyzed MixUp\cite{ref16}, CutMix\cite{ref32}, and FMix\cite{ref39} from an information theoretic perspective. In mixup, however, it remains mysterious why the output of a linear mixture of two training samples should be the same linear mixture of their labels(especially in highly nonlinear models)\cite{ref25}. And Guo et al.\cite{ref29} provide an example of how mixed labels can conflict with actual data point labels. Chidambaram et al.\cite{ref25} analyzed that the advantages of mixup depend on the properties of the data. Han et al.\cite{ref48} analyzed that the failure of the mixup may be that the synthesized data points are still soft-connected to the original labels, and propose an Universum-style mixup that is disconnected from all known class labels.

However, simply combining traditional mixup(low-order mixup \emph{i.e.} mixing only once) and Universum is not suitable for imbalanced scenarios.  In this paper, we are the first to propose a high-order mixup paradigm to generate a CaU that can avoid the constraints of the original label combination. This CaU not only can effectively solve the long-tailed distribution, but also has universality to other small-sample scenarios.
\begin{figure*}
\begin{center}
   \includegraphics[scale=0.60]{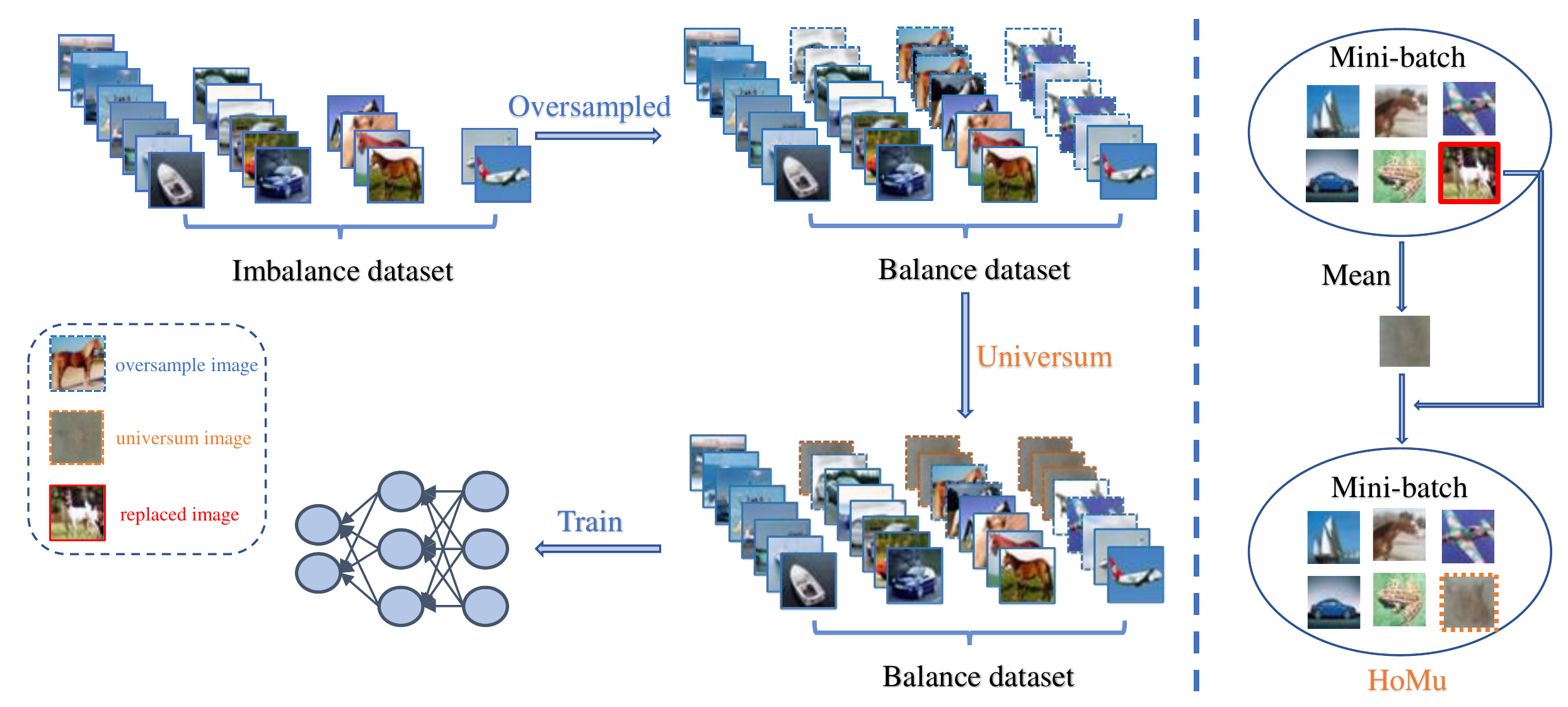}
   \caption{\emph{\textbf{\emph{Model overview. (Left)}}} CaUIRL performs re-balance learning by first balancing the long-tailed data using over-sampling and replacing the over-sampled data with Universum data  by probability, so CaUIRL uses both the over-sampled data and the Universum data. \textbf{(Right)} The right figure shows the HoMu operation process. First, average the pictures in a mini-batch, and then add the replaced pictures to the mini-batch as Universum. }
   \label{3}
\end{center}
\end{figure*}
\section{Analysis for Long-Tailed Recognition}

\subsection{Problem Setup}
For a given unbalanced training dataset $\mathcal{D}=\left\{\left(x_{i}, y_{i}\right)\right\}_{i=1}^{N}$, where $x \in \mathbb{R}^{d}$ and
$y \in\{1,2, \ldots, C\}$, each class $y_{i}$ consists of $N_{i}$ training samples. Let $N:=\sum_{i} N_{i}$ denote the total number of samples in $\mathcal{D}$. Without losing generality, we assume that $N_{1} \geq N_{2} \geq \ldots \geq N_{C}$, then the imbalance rate can be defined as $v=\frac{N_{max}}{N_{min}}$($max=1$, $min=C$), and the test set is balanced. We define $z=f(x ; \theta)$ as the feature of sample $x$, where $f: \mathcal{X} \rightarrow \mathcal{Z}$ projects the input into the feature space $\mathcal{Z} \subseteq \mathbb{R}^{s}$. The final prediction is $\hat{y}=g(z)$,$\quad g: \mathcal{Z} \rightarrow \mathcal{Y}$ is the classification network. Let $P_{train}(X, Y)$, $P_{test}(X, Y)$ be defined as the underlying training and test distribution. Generally, the class imbalance problem assumes that the test data has the same class conditional probability as the training data, \emph{i.e.}, $P_{train}(X \mid Y)=P_{test}(X \mid Y)$, while their class priors are different, \emph{i.e.}, $P_{train }(Y) \neq P_{test }(Y)$. It is obtained by the multiplication formula, $P_{train }(X,Y) \neq P_{test }(X,Y)$.
Therefore, the primary purpose of class imbalance classification is to use the training set $\mathcal{D} \sim \mathcal{P}_{train }(X, Y)$ to train $f$ and $g$, which can have better generalization on the test set.
\subsection{Problem Analysis}
In this section, we explain the problem of long-tailed recognition from two levels of gradient and domain shift.
\subsubsection{Gradient Level}
As shown in Fig.\ref{1}(a), when learning a classification model in long-tailed data, tail classes are easily overwhelmed by head classes\cite{ref10}\cite{ref55}. Below we elaborate on the factors underlying this issue from a gradient perspective.

Without loss of generality, taking binary classification as an example, we train a general classifier by gradient on  the cross entropy:

\begin{equation}\label{eqn:1}
\begin{aligned}
\mathcal{L}_{C E}(\Theta) &:=\frac{1}{N} \sum_{i=1}^{N}-\left[y_{i} \log p_{i}+\left(1-y_{i}\right) \log \left(1-p_{i}\right)\right] \\
                          &:=\frac{1}{N} \sum_{i=1}^{N} l_{i, C E}\left(x_{i}, \Theta\right) \\
                          &:=\frac{1}{N_{1}} \sum_{i=1}^{N_{1}} l_{i, C E}\left(x_{i}, \Theta\right)+\frac{1}{N_{2}} \sum_{i=N_{1}+1}^{N} l_{i, C E}\left(x_{i}, \Theta\right)
\end{aligned}
\end{equation}
where $p=g(f(x, \Theta))$, $N_{i}$ is the number of samples of the $i$-th class.

The gradient of $l_{CE}$ with respect to the model parameters $\Theta$ is
\begin{equation}
\nabla l_{C E}(x, \Theta)=\nabla g(f(x, \Theta))(g(f(x, \Theta))-y).
\end{equation}

The gradient of the cross-entropy loss with respect to $\Theta$ satisfies
\begin{equation}
\nabla L_{C E}(\Theta)=\nabla L_{C E, N_{1}}(\Theta)+\nabla L_{C E, N_{2}}(\Theta), \nonumber
\end{equation}
where $\nabla L_{C E, N_{k}}(\Theta)=\frac{1}{N_{k}} \sum_{i=1}^{N_{k}} \nabla l_{i, C E}\left(x_{i}, \Theta\right), k=(1,2)$.
The gradient of layer $l$ of the model is updated as
\begin{equation}
w_{l}^{t+1}=w_{l}^{t}-\eta\left(\nabla L_{C E, N_{1}}(\Theta)+\nabla L_{C E, N_{2}}(\Theta)\right).
\end{equation}
According to equation (3), it can be observed that gradient update are determined by $(\nabla L_{C E, N_{1}}(\Theta)+\nabla L_{C E, N_{2}}(\Theta))$. Therefore,  the influence of minority classes on the gradient is easily ignored when $N_{1} \gg N_{2}$.

According to formula (2), we find that the loss is consistent with the gradient direction of the predicted value, and the difference between the predicted value and ground truth is regarded as the weight. Further analysis shows that although the magnitudes of the gradient components of the minority classes increase under traditional over-sampling, their directions will remain relatively unchanged due to the lack of diversity. As a result, the traditional over-sampling limited the generalization of minority classes.

\subsubsection{Domain Shift Level}
In long-tailed recognition, the general setting is that the training set is unbalanced and the test set is balanced. Such a setup inevitably incurs minority classes domain shift between training and test sets, since the insufficient samples of minority classes make their expression biased. To reflect such shift, following the nonparametric Maximum Mean Discrepancy (MMD)\cite{ref56}, we here provide a measure of the \emph{corresponding class gap (C2G)} in \emph{Definition1} to approximately cater for the scenario. However, unlike MMD used in transfer learning and domain adaptation to empirically estimate the discrepancy between dataset distributions\cite{ref57},\cite{ref58},\cite{ref59}, C2G is introduced to roughly measure the gap between individual classes in different datasets (\emph{i.e.}, training and test datasets). And in subsection 4.2, through experiments we further find that the appropriate Universum can be effectively selected with the help of C2G.

~\\
{
\setlength{\parindent}{0cm}
\textbf{Definition 1.}  \emph{(Corresponding Class Gap(C2G)) For two different datasets $\mathcal{D}_{1}$ ,$\mathcal{D}_{2}$ from the same label space $\mathcal{C}$, their feature spaces are denoted as $\mathcal{Z}_{1}$ , $\mathcal{Z}_{2}$, respectively. Given a distance function $d: \mathbb{R}^{s} \times \mathbb{R}^{s} \rightarrow \mathbb{R}$ in the feature space, for $\forall c \in \mathcal{C}$, the C2G between different datasets is defined as,}
\begin{equation}
\operatorname{gap}(c)=d\left(\mu_{\mathcal{Z}_{1}, c}, \mu_{\mathcal{Z}_{2}, c}\right), \mu_{\mathcal{Z}, c}=\mathbb{E}_{{z \in \mathcal{Z}}, {i \in c}}[z_{i}],
\end{equation}

where $d$ defaults to Euclidean distance, $\mu_{\mathcal{Z}, c}$ represents the mean of the \emph{c-th} sample in the feature space $\mathcal{Z}$ . In general, more severe domain shift is accompanied by larger gaps, and vice versa. }

To verify that C2G can reflect such a domain shift, we perform the following experiments. Concretely, we set $v=50$ and $v=100$ to construct the long-tailed dataset Cifar-10-LT, respectively, and the label distribution is shown in Fig.\ref{2}(a). We train a vanilla feature extractor using the WideResNet-28-10 whose loss function is the cross-entropy function. We calculate the C2G between the training and the test set according to \emph{Definition1}, and the final result is shown in Fig.\ref{2}(b).

In Fig.\ref{2}(b), we find that the C2G value gradually increases as the number of intra-class samples in the training set decreases. This indicates that as the C2G value between the training and the test set increases, their corresponding classes are pulled farther apart in the feature space, resulting in a more pronounced domain shift.
In addition, we observe that C2G grows with rising imbalance rate $v$, suggesting that domain shift becomes more severe with increasing $v$. In summary, as the number of intra-class samples decreases, there is more severe domain shift, which can be roughly measured by simply calculating the C2G value.

\section{Class-Aware Universum for Long-Tailed Recognition}
This section introduces our algorithm and gives theoretical guarantees from a Bayesian perspective.
\subsection{Method Introduction}
Different from existing methods, we here borrow Universum to realize the sample re-balance of minority classes. Note that such a borrowing is not-trivial since directly using Universum as a whole contradictory class like the traditional way will fall into the dilemma of OPeN mentioned above. To overcome this, with the help of Universum's own discrimination, we innovatively endow Universum with class-aware capability, where we re-balance the minority classes by adding certain category specific Universum.
\begin{table*}[t]
    \centering
    \caption{C2G values corresponding to different Universum datasets.}
    \label{tab:univ-compa}
    \begin{tabular}{lccccccccccr}
    \toprule
        \textbf{Dataset} & \textbf{0} & \textbf{1} & \textbf{2} & \textbf{3} & \textbf{4} & \textbf{5} & \textbf{6} & \textbf{7} & \textbf{8} & \textbf{9} & \textbf{Mean} \\ \midrule
        U-IMG & 45.48 & 73.18 & 39.21 & 37.08 & 46.79 & 34.94 & 42.56 & 45.91 & 48.7 & 51.65 & 45.92 \\
        U-TINMG & 30.21 & 69.17 & 35.06 & 28.28 & 38.88 & 34.24 & 39.65 & 39.12 & 46.22 & 48.34 & 40.91 \\
        U-SVHN & 17.34 & 54.13 & 22.86 & 15.01 & 23.35 & 18.88 & 25.74 & 29.20 & 35.50 & 34.76 & 27.68 \\
g-Universum & \textbf{14.99} & \textbf{30.82} & \textbf{14.84} & \textbf{4.72} & \textbf{10.27} & \textbf{12.23} & \textbf{18.53} & \textbf{14.63} & \textbf{20.26} & \textbf{28.01} & \textbf{16.93} \\
    \bottomrule
    \end{tabular}
\end{table*}

Concretely, for a given unbalanced dataset $\mathcal{D}$,  we define the sampling probability of each class as $p_{c}=\frac{1}{N_{c}}(c=1,...,C)$, then we randomly sample the dataset with weights according to $p_{c}$ to obtain an over-sampled balanced dataset. Constructing Universum dataset $\mathcal{D}_{u m i}=\left\{\left(\tilde{x}_{i}, y_{i}\right)\right\}_{i=1}^{N_{uni}}$, where $\tilde{x} \in \mathbb{R}^{d}$ and $y \in\{1,2, \ldots, C\}$. $\mathcal{D}_{u m i}$ can be either external data or generated from raw data, as discussed in detail in subsection 5.1. According to \cite{ref10}, we define the \emph{representation-rate} of each class as $p_{r_{c}}=\frac{N_{c}}{N_{\max }}(c=1,...,C)$, then the probability that the $i$-th sample of the $c$-th class is replaced by the Universum is,
\begin{equation}
p_{u_{c}}^{i}=\left(1-p_{r_{c}}\right) \delta,
\end{equation}
where $\delta \in[0,1]$ is used to balance the replacement rate of Universum and the default value of $0.9$. It is not difficult to find from equation (5) that a lower $p_{r_{c}}$ leads to a higher probability of replacing samples of class $c$ with the Universum.

Finally, the samples replaced by Universum are selected according to the Bernoulli distribution, that is,
\begin{equation}
\operatorname{\emph{Bernoulli}}\left(p_{u_{c}^{i}}\right)=\left\{\begin{array}{lr}
1, & \text { \emph{$x_{i}^{c} \leftarrow \tilde{x}_{j}^{c}$} } \\
0, & \text { \emph{unreplaced} }
\end{array}\right.
\end{equation}
where $\tilde{x}_{j}^{c}$ represents a random sample drawn from \emph{c-th} class in $\mathcal{D}_{u m i}$. According to formula (6), it is not difficult to find that we add Universum to the given target class, and assign the label of the target class to Universum. This kind of Universum is referred to as a class-aware Universum(CaU). The overall framework of our algorithm is shown on the left of Fig.\ref{3}.

\subsection{Theoretical analysis}
In long-tailed recognition, due to the different class prior probabilities of the training and test sets, the joint probabilities of the two are also inconsistent, that is,
\begin{equation}
P_{train }(X,Y) \neq P_{test }(X,Y)
\end{equation}
equivalent to
\begin{equation}
\underset{y \in \mathcal{Y}}{\arg \max}\ p_{{train}}(x\!\mid y) p_{{train}}(y)\neq \underset{y \in \mathcal{Y}}{\arg \max }\ p_{{test }}(x \mid y) p_{{test }}(y).
\end{equation}
The above formula implies that it is unreasonable to directly apply the Bayesian classifier learned on the training set to the test set, which will result in a reduction in the generalization performance of the model. This is also the primary cause behind the poor performance of the long-tailed recognition.

We here provide the theoretical evidence that the classifiers learned by CaUIRL are consistent with those learned under the balanced condition from a Bayesian perspective.

~\\
{
\setlength{\parindent}{0cm}
 \textbf{Theorem 1.}}  \emph{Let $p_{{train }}(x \mid y)$ and $p_{{uni }}(\tilde{x} \mid y)$ be the class conditional probabilities of the training set and Universum set, respectively, if $p_{{train }}(x \mid y)=p_{{uni }}(\tilde{x} \mid y)$ holds, then we have}
\begin{equation}
\underset{y \in \mathcal{Y}}{\arg \max}\ p_{{re}}(x\mid y) p_{{re}}(y)=\underset{y \in \mathcal{Y}}{\arg \max }\ p_{{test }}(x \mid y) p_{{test }}(y),
\end{equation}
\emph{where $p_{\text {re}}(x\mid y)$ and $p_{{test}}(x \mid y)$ represent the class conditional probabilities of the training set  after balancing and test set.}
~\\

The proof is provided in Appendix B of the supplementary material. According to formula (6), the Universum is added according to the category of the target data, that is, the Universum is class-aware in CaUIRL, then we have
\begin{equation}
p_{{train }}(x| y)=p_{{uni }}(\tilde{x} | y)
\end{equation}
It can be known from Theorem 1 that Equation (10) is a sufficient condition for Equation (9) to hold.

Therefore, from the perspective of Bayesian theory, it can be seen that the classifier obtained by CaUIRL is consistent with the balanced scene, which provides a theoretical guarantee for our algorithm.

\section{What Is Good Universum in Long-tailed Recognition}
\subsection{The effect of domain similarity}

In CaUIRL, we re-balance all categories of data by adding Universum to the corresponding target data. However, there may be many candidates for the Universum used in CaUIRL, and different Universum datasets may bring different degrees of domain shift to the model. Therefore, this subsection utilizes our proposed C2G to explore the impact of different Universum datasets.

\subsubsection{External Universum for CaUIRL}
Universum generally refers to one or more external datasets independent of the target dataset, and we here call such datasets external Universum. Next, we specifically explore the impact of different external Universum datasets on the model.

Two baseline methods: (1) Empirical Risk Minimization (ERM): Training without any re-balancing scheme. (2) Over-sampling: Re-balancing the dataset by over-sampling minority classes with augmentations. In addition, we also use different external Universum datasets to build models, denoted as U-IMG, U-TINMG and U-SVHN(Appendix A of the supplementary material presents the relevant Universum datasets). We employ Cirar-10-LT ($v=50$ and $v=100$) as the imbalanced dataset, WideResNet-28-10 network as the backbone, and cross-entropy as the loss.

As demonstrated in Fig.\ref{2}(c), the classification results of the algorithms employing external Universum (U-IMG, U-TINMG, U-SVHN) are greatly improved as compared to the baselines (over-sample, ERM), indicating that using external Universum directly in CaUIRL is effective for long-tailed issue. Specifically, at $v=50$ and $v=100$, U-SVHN achieves the best classification accuracy, which is $4.6\%$ and $8.66\%$ higher than over-sampling. Furthermore, we find that using different Universum datasets corresponds to different model performance. This indicates that the quality of the Universum itself directly impacts model performance. What kind of Universum data is the most effective? With this in mind, we further calculated the C2G values of the corresponding classes between different Universum datasets and Cifar-10-LT, and the results are shown in Table 1.

It can be seen from Table 1 that the average C2G values are SVHN, TinyImagenet10 and Imagenet10 from high to low, while the classification accuracy of the corresponding models is in the opposite order. It shows that the model with the highest classification accuracy is the model with the smallest C2G average. This is because the C2G value measures the degree of domain shift, and when the C2G mean is smaller, the domain shift is slighter. So in order to reduce the impact of domain shift on model performance, we hope that the introduced Universum has a smaller C2G.

\subsubsection{Auto-generated Universum for CaUIRL}
From subsection 5.1.1, we can find that class-aware Universum indeed enhances the performance of long-tailed recognition. In particular, the smaller the C2G value of the Universum, the better the performance obtained. More often than not, however, we may have some Universum datasets, but the C2G value between them and the target data may be very large, or even more, we may not have Universum available at hand. To address these challenges, we innovatively propose an auto-generated Universum as CaU based on a popular mixup\cite{ref16} strategy without resorting to any external data. Below we describe its generation process in detail.

Zhang et al.\cite{ref16} proposed the mixup method, a data agnostic and straightforward data augmentation principle. Mixup is a form of vicinal risk minimization that generates virtual samples by sampling two random samples of the training set and its labels, and linearly interpolating them. The main idea of mixup is that for two randomly sampled pairs $\left\{x_{i}, y_{i}\right\}$ and $\left\{x_{j}, y_{j}\right\}$, with:
\begin{equation}
\begin{split}
x=\lambda x_{i}&+(1-\lambda) x_{j}, \\
y=\lambda g\left(f\left(x_{i}\right)\right)&+(1-\lambda) g\left(f\left(x_{j}\right)\right)
\end{split}
\end{equation}
where $\lambda \in \operatorname{Beta}(\alpha, \alpha)$, for $\alpha \in(0, \infty)$. As a data augmentation technique, mixup can significantly improve generalization error of the model. However, affected by label mixing, we find that (11) holds if and only if $f$ and $g$ are linear, which is a challenging condition for deep models.

\begin{figure}
\begin{center}
   \includegraphics[scale=0.5]{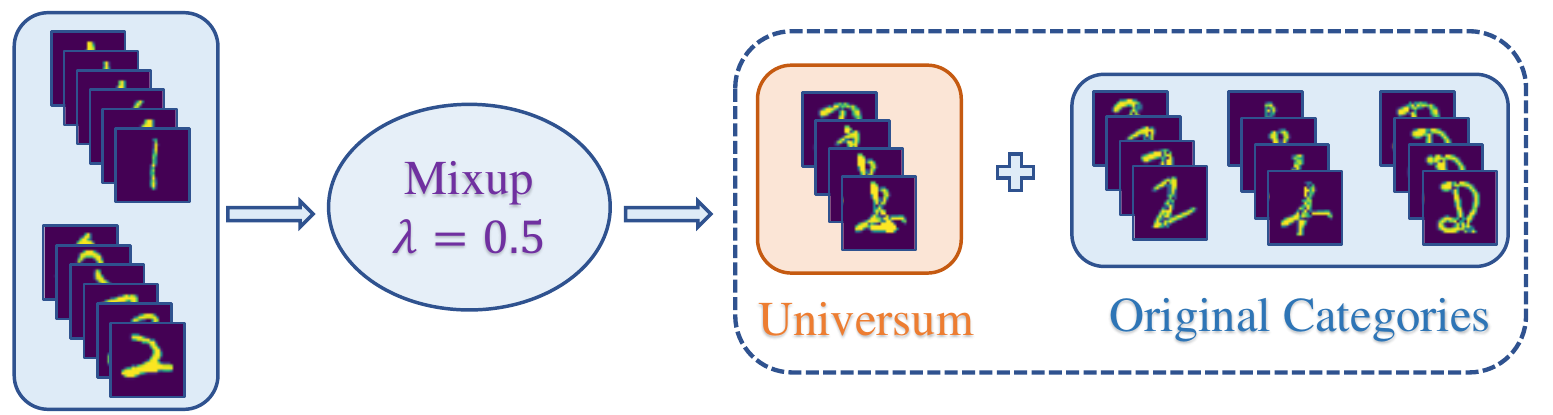}
   \caption{\textbf{Mixup generates pictures.} Input the handwritten digits "1" and "2" , then using the mixup function ($\lambda$=0.5) to possibly generate Universum data (orange rectangle) and data belonging to the original categories (blue rectangle).}
   \label{4}
\end{center}
\end{figure}

To avoid the effect of label combination, Han et al.\cite{ref48} used the images generated by only using  sample-level mixup operation as Universum for supervised contrastive learning, and the model performance was significantly improved. However, this operation may violate the definition of Universum.
For example, considering the following toy example, for the MINIST dataset, a mix of the number "1" and the number "2" might yield a class that belongs to the original dataset. As shown in Fig.\ref{4}, directly used for mixed data, the generated pictures may be included in categories such as "2", "4", "8", etc. To address the above problems, we propose \emph{Higher-Order Mixup(HoMu)} operations to automatically  generate Universum. The specific operation is for a given batch of dataset $\mathcal{B}=\left\{x_{1}, x_{2}, \cdots, x_{n}\right\}$, for sample $x_{i} \in \mathcal{B}$, its\emph{ High-Order Mixup} is,

\begin{equation}
\tilde{x}_{i}=(1-\lambda) \frac{1}{|\mathcal{B}|} \sum_{j=1}^{n} x_{j}+\lambda x_{i}.
\end{equation}
where $\lambda$ is recorded as the mixing coefficient. The right side of Fig.\ref{3} is our operation process. We record this type of Universum(\emph{i.e.}, $\tilde{x}_{i}$) as \emph{g-Universum}, which satisfies $p\left(x_{i} \mid y_{i}\right)=p\left(\tilde{x}_{i} \mid y_{i}\right)$. Compared with the traditional mixup operation, the HoMu operation is more thoroughly mixed, which can avoid the generated data belonging to the original category. According to Table 1, \emph{g-Universum} has the smallest C2G value compared to other types of Universum. This shows that using \emph{g-Universum} can significantly reduce the gap with the target dataset, which can further alleviate the domain shift.

\subsection{The effect of class separability}

\begin{figure}
\begin{center}
   \includegraphics[scale=0.37]{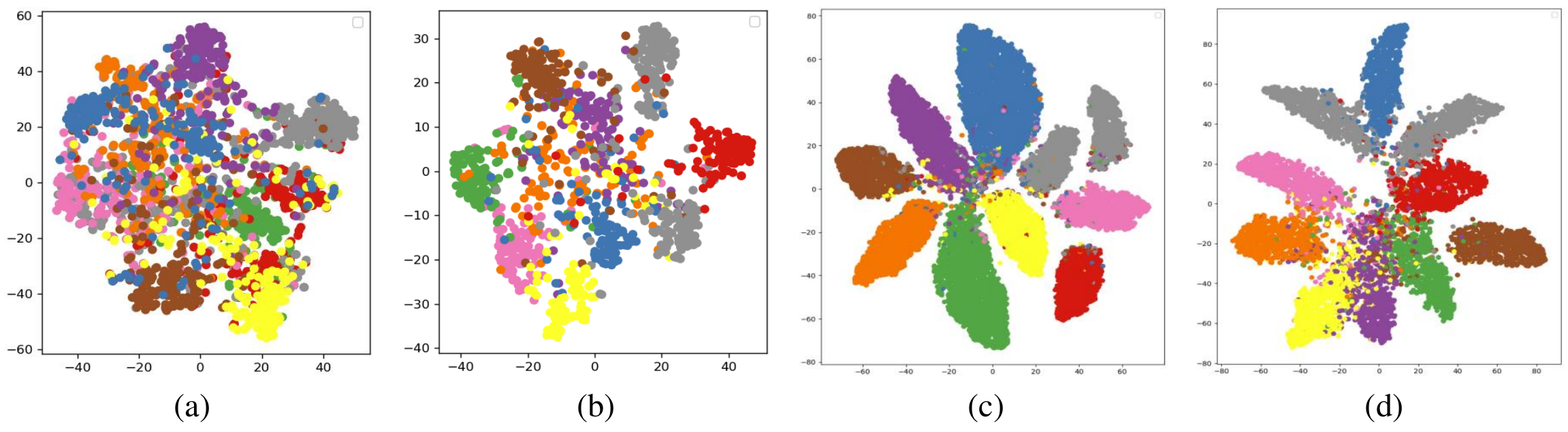}
   \caption{\textbf{Different Universum datasets clustering effect legends.}(a)-(d) correspond to the use of different Universum datasets in CaUIRL (U-IMG, U-
TINMG, U-SVHN, \emph{g-Universum}) clustering results.}
   \label{cluster}
\end{center}
\end{figure}

In our method, we endow the Universum with class-aware ability to re-balance individual minority classes, thus the separability of the Universum itself may also affect the performance of the model. In this subsection, we explore this effect using the visualized clustering results. Specifically, we utilize the Resnet18 to train 100 epochs , then the \emph{k}-means algorithm to cluster different Universum datasets (U-IMG, U-TINMG, U-SVHN, \emph{g-Universum}), and use t-SNE(t-distributed Stochastic Neighbor Embedding)\cite{ref47} to reduce the features for visualization, and the final result is shown in Fig.\ref{cluster}.

Combining Fig.\ref{relationship} and (a)-(c) of Fig.\ref{cluster}, we observe that the (c) has better separability compared to the (a) and (b), and compared to the U-IMG and U-TINMG, U-SVHN has higher accuracy. This reflects that the better the separability of the used Universum itself, the better the corresponding model performance. However, we also find that (d) is less separable but has better recognition performance compared to (c). This shows that when using Universum in CaUIRL, we cannot blindly pursue the separability of the Universum itself, but also consider the impact of domain shift caused by adding external data.

\subsection{The effect of sample diversity}
\begin{figure}
\begin{center}
   \includegraphics[scale=0.38]{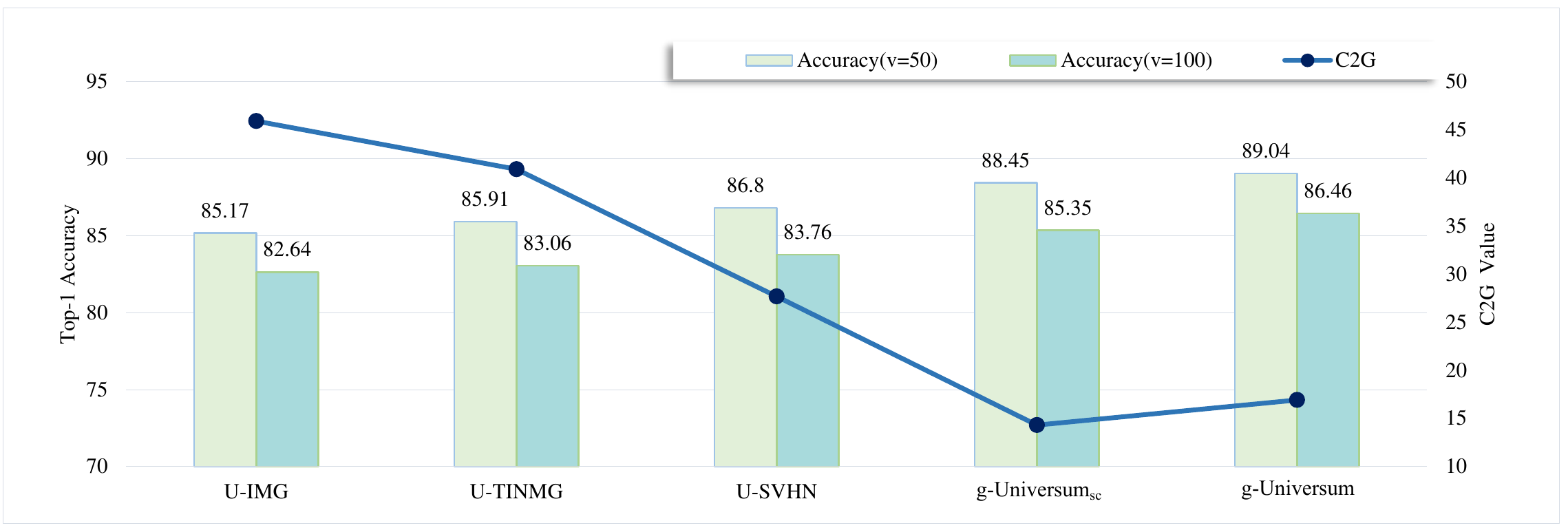}
   \caption{\textbf{The relationship between accuracy and C2G value.}}
   \label{relationship}
\end{center}
\end{figure}

Sample diversity is also a key factor in data augmentation methods for long-tailed identification. To explore the impact of sample diversity brought by using HoMu auto-generated Universum on model performance, in this subsection, we reconstruct a new strategy to automatically generate Universum using only the samples of the same class(refer to Appendix A of the supplementary material), referred to as \emph{g-Universu$m_{sc}$}. We calculate the C2G values and corresponding model top-1 accuracies of five different Universum datasets(U-IMG, U-TINMG, U-SVHN, \emph{g-Universu$m_{sc}$}, \emph{g-Universum}) separately, as shown in Fig.\ref{relationship}.

We can find that for different imbalance rates, the model using \emph{g-Universum} achieves the best accuracy, followed by the model using \emph{g-Universu$m_{sc}$}. It is shown that only using the automatically generated Universum from the target data is more suitable for long-tailed recognition, because using the automatically generated Universum to avoid the introduction of external data can alleviate the problem of domain shift. However, we find that \emph{g-Universu$m_{sc}$} has the smallest C2G value, but the second highest accuracy, which seems to be contrary to our previous discussion. This is because \emph{g-Universu$m_{sc}$} is generated using only samples from the same class, while \emph{g-Universum} is generated using data from all classes in the mini-batch. Compared with \emph{g-Universum}, \emph{g-Universu$m_{sc}$} lacks sample information from other classes. Therefore, although the C2G value corresponding to \emph{g-Universu$m_{sc}$} is lower than \emph{g-Universum}, the model performance corresponding to \emph{g-Universu$m_{sc}$} is suboptimal due to the lack of sample diversity.

~\\
{
\setlength{\parindent}{0cm}
\textbf{Remark.}  Based on the above experimental analysis, we can obtain that domain similarity, class distinction and sample diversity should be considered when selecting Universum for long-tailed recognition. And according to the experiments, we also find that the Universum proposed in this paper, which is automatically generated according to HoMu, has all three properties at the same time. \emph{g-Universum} is used in all subsequent experiments if the model is not specifically stated.
}The pseudo-code of CaUIRL is shown in Algorithm 1.

\renewcommand{\algorithmicrequire}{ \textbf{Input:}}     
\renewcommand{\algorithmicensure}{ \textbf{Output:}}    
\begin{algorithm}
\caption{CaUIRL for Long-Tailed Recognition}
\label{alg1}
\begin{algorithmic}[1]
\Require Imbalanced dataset $\mathcal{D}$,  parameter $\delta, \lambda\in[0,1]$
\Ensure Mini-Batch $\mathcal{B}$.
\State Calculate sampling probability $\{p_{s_{1}}, p_{s_{2}},...,p_{s_{C}}\}$. \\
$\mathcal{D}_{re}$  $\leftarrow$ Constructing over-sampling balanced dataset.
\While{$t<=MaxEpoch$} \\
\ \ \ \ \ $\mathcal{B}=\left\{\left(x_{i}, y_{i}\right)\right\}_{j=1}^{batch\_size }$ $\leftarrow$ Sample a batch from $\mathcal{D}_{re}$
\For
{$\{x_{i}, y_{i}\}$ in $\mathcal{B}$}\\
\ \ \ \ \ \ \ \ \ \ \ $p_{u_{i}} \leftarrow \left(1-p_{r_{i}}\right) \delta$\\
\ \ \ \ \ \ \ \ \ \ \ $b_{u_{i}} \leftarrow Bernoulli(p_{u_{i}})$
\If{$b_{u_{i}} == 1$}
\State $x_{uni} \gets (1-\lambda) \frac{1}{|\mathcal{B}|} \sum x_{j}+\lambda x_{i}$
\State $x_{i} \gets x_{uni}$
\EndIf

\EndFor
\\
\ \ \ \ \ \ \textbf{return} $\mathcal{B}$
\EndWhile

\end{algorithmic}
\end{algorithm}

\section{Experiments}
\subsection{Datasets and Setup}
Our experimental setup (including implementation details and evaluation protocol) mainly follows\cite{ref12} for Cifar-10-LT and Cifar-100-LT, and \cite{ref41} for CelebA-5. We follow the evaluation protocol used in\cite{ref4}\cite{ref10}\cite{ref11} for imbalanced classification tasks.


\subsubsection{Datasets Explanation}
We use hand-crafted imbalanced datasets (Cifar-10-LT, Cifar-100-LT) and real-world imbalanced datasets (CelebA-5) in our experiments, which are detailed below.

\textbf{Cifar-10-LT and Cifar-100-LT}\cite{ref1}. Cifar-10 and Cifar-100 both have 60,000 images, 50,000 for training and 10,000 for test with 10 categories and 100 categories.  For fair comparison, we use the long-tailed versions of Cifar datasets with the same setting as those used in a\cite{ref10}. It builds the long-tailed version Cifar-10/100-LT(see Table\ref{tab:Datasets} for statistics) by sampling the number of images per class in the training set with exponential decay, while their corresponding test set remains unchanged (\emph{i.e.} uniform class distribution). Correspondingly, we also evaluate our method on challenging dataset settings (v=50, v=100).

\textbf{CelebA-5}\cite{ref42}. CelebA was originally a multi-label dataset consisting of face images, the CelebA-5(see Table\ref{tab:Datasets} for statistics) dataset was proposed in a\cite{ref43} by selecting from non-overlapping attributes of hair color (blond, black, bald, brown, gray) sample. Naturally, the resulting dataset is unbalanced ($v = 10.7$) because the color of human hair is not uniformly distributed. Then resize the image to 64x64 pixels. Reference a\cite{ref41} also subsampled the dataset by 1/20 while maintaining the imbalance ratio $v$ to make the task more difficult.
\begin{table*}[t]
    \centering
    \caption{Statistics of the Three Datasets Utilised in Our Evaluations}
    \label{tab:Datasets}
    \begin{tabular}{lccccccc}
    \toprule
        \textbf{Dataset} & \textbf{Semantics} & \textbf{Classes} & \textbf{Pixels} & \textbf{Imbalance Rate} & \textbf{Largest Class Size} & \textbf{Smallest Class Size} & \textbf{Total Images}  \\ \midrule
        Cifar-10-LT & Object Category & $10$ & $32\times32$& $v=50/100$ & $5000$ & $\{100,50\}$ & $\{13996, 12406\}$  \\
        Cifar-100-LT & Object Category & $100$ & $32\times32$& $v=50/100$ & $500$ & $\{10,5\}$ & $\{12408, 10847\}$  \\
        CelebA-5 & Facial Attribute & $5$  & $64\times64$& $v=50/100$ & $2423$ & $227$ & $6651$  \\
    \bottomrule
    \end{tabular}
\end{table*}

\begin{table*}[t]
  \centering
  \caption{Results on Cifar-10-LT, Cifar-100-LT, CelebA-5.}
	\begin{threeparttable}


    \begin{tabular}{cc|ccccccc}
    \toprule
    \multicolumn{2}{c|}{\textbf{Methods}} & \textbf{ERM}   & \multicolumn{1}{p{7.665em}}{\textbf{Over-sampling}} & \multicolumn{1}{p{6.945em}}{\textbf{LDAM-DRW}} & \multicolumn{1}{p{4.055em}}{\textbf{M2m}} & \multicolumn{1}{p{4.055em}}{\textbf{MiSLAS{*}}} & \multicolumn{1}{p{4.055em}}{\textbf{OPeN}} & \multicolumn{1}{p{4.055em}}{\textbf{CaUIRL}} \\
    \midrule
    \multicolumn{1}{c}{\multirow{2}[1]{*}{Cifar-10-LT}} & \multicolumn{1}{p{4.055em}|}{v=100} & 79.6(0.2) & 75.1(0.4) & 80.5(0.6) & 81.3(0.4) & 82.1  & \underline{84.6(0.2)} & \textbf{86.46(0.2)} \\
          & \multicolumn{1}{p{4.055em}|}{v=50} & 84.9(0.4) & 82.2(0.4) & 85.3(0.2) & 85.5(0.3) & 85.7  & \underline{87.9(0.2)} & \textbf{89.04(0.1)} \\
    \multicolumn{1}{c}{\multirow{2}[0]{*}{Cifar-100-LT}} & \multicolumn{1}{p{4.055em}|}{v=100} & 47.0(0.5) & 42.5(0.3) & 46.8(0.2) & 46.5(0.5) & 47    & \underline{51.5(0.4)} & \textbf{52.64(0.2)} \\
          & \multicolumn{1}{p{4.055em}|}{v=50}  & 52.4(0.4) & 48.0(0.2) & 52.6(0.2) & 52.9(0.2) & 52.3  & \underline{56.3(0.4)} & \textbf{57.35(0.1)} \\
    CelebA-5 & \multicolumn{1}{p{4.055em}|}{v=10.7} & 78.6(0.1) & 76.4(0.2) & 78.5(0.5) & 76.9(0.4) & -     & \underline{79.7(0.2)} & \textbf{81.48(0.3)} \\
    \bottomrule
    \end{tabular}%
\begin{tablenotes}
			\footnotesize
			\item \emph{The experimental results in the table are consistent with those in \cite{ref10}. {*} indicates the results in the original paper, the others are results obtained with the same settings as our model. Missing results indicate datasets not evaluated in the cited papers.}
			
		\end{tablenotes}
	\end{threeparttable}

  \label{accuracy}%
\end{table*}%
\subsubsection{Experimental Setup}
Without special instructions, all experiments in this subsection use WideResNet-28-10\cite{ref44}, where DAR-BN\cite{ref10} is used to replace the original BN layer in order to alleviate the problem of domain shift. For Cifar-10/100-LT, we follow the settings in \cite{ref10} and \cite{ref41},\emph{ i.e.} training $200$ SGD optimizers with cross-entropy loss using a momentum of $0.9$ and weight decay epoch $2e-4$. We set the mini-batch size to $128$, use a step learning rate decay with an initial learning rate of $0.1$, and then decay by a factor of $0.01$ at epochs $160$ and $180$. We additionally choose a linear warm-up learning rate technique for the first $5$ epochs to be compatible with some baseline methods, even though it has minimal bearing on our method \cite{ref45}. CaUIRL is deferred to  the last $40$ epochs. For CelebA-5, we train for $90$ epochs and reduce the learning rate by a factor of $0.1$ at epochs $30$ and $60$, consistent with \cite{ref42}. In the experiments, for CelebA-5, Cifar-10-LT and Cifar-100-LT, we repeated the experiment 5 times and reported the mean and standard error.

In each experiment, our data augmentation principle based on \cite{ref10} reduces the random GaussianBlur operation as follows: For the Cifar-10/100-LT and CelebA-5 datasets, we use random horizontal flip followed by random crop with padding of four pixels, and then applied color dithering, random grayscale, and Cutout\cite{ref46}.

\subsection{Performance Comparison}
\subsubsection{Compared methods}
We compared our approach with the following methods. \textbf{\emph{(A) Baselines:}} empirical risk minimization (ERM): training on the cross-entropy loss without any re-balancing; Oversampling: re-balancing the dataset by oversampling minority classes with augmentations. These baseline methods share the same training parameters, augmentations, and architectures as our algorithm. \textbf{\emph{(B)Recent Leading Methods:}} label-distribution-aware margin(deferred re-weighting, LDAM-DRW)\cite{ref12}£ºthe classifier is trained to impose larger margin to minority classes; major-to-minor translation(M2m)\cite{ref41}: augment minority classes by augmenting samples from majority classes; mixup shifted label-aware smoothing(MiSLAS)\cite{ref13}: they propose label-aware smoothing to deal with different degrees of over-confidence for classes; oversampling with pure noise images(OPeN)\cite{ref10}:using pure noise images as additional training data for long-tailed recognition, where the OPeN is the current SOTA method. Table\ref{accuracy} shows the performance of these algorithms and our algorithm for long-tailed recognition.

\begin{figure}
\begin{center}
   \includegraphics[scale=0.60]{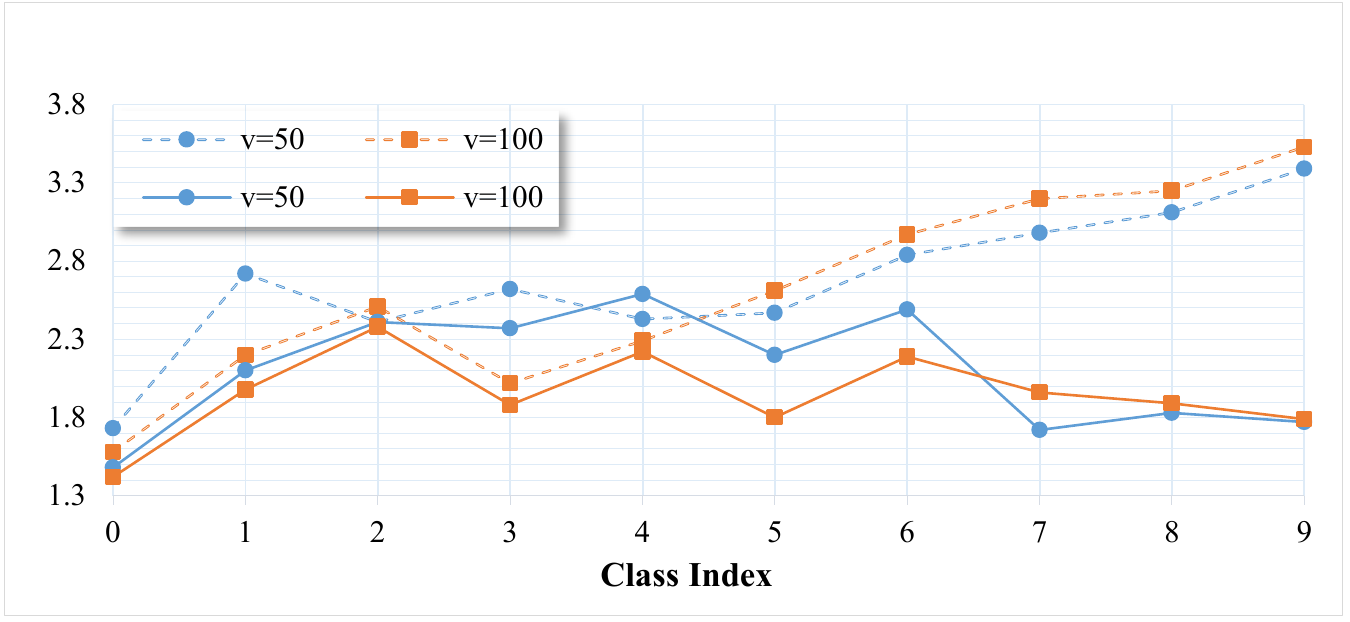}
   \caption{\textbf{C2G value comparison.} The dotted line represents the C2G value calculated with ERM, and the solid line represents the C2G value after using CaUIRL.}
   \label{rebalance}
\end{center}
\end{figure}

\subsubsection{Experimental results}
According to Table\ref{accuracy}, it can be seen that our algorithm is consistently higher than other algorithms on different datasets. Among them, when $v=100$ on the Cifar-10-LT dataset, CaUIRL has the best recognition effect compared to other algorithms, which is $6.86\%$ more accurate than ERM and $1.86\%$ higher than OPeN. When $v=50$, the overall accuracy of CAUIRL on Cifar-100-LT has the smallest advantage over other algorithms, but it is still $4.95\%$ higher than ERM algorithm and $1.05\%$ higher than OPeN. In summary, the CaUIRL algorithm can further improve the performance of long-tailed recognition on different datasets and achieve SOTA results. This shows that it is effective to use the automatically generated CaU for rebalancing learning.

\begin{figure*}
\begin{center}
   \includegraphics[scale=0.530]{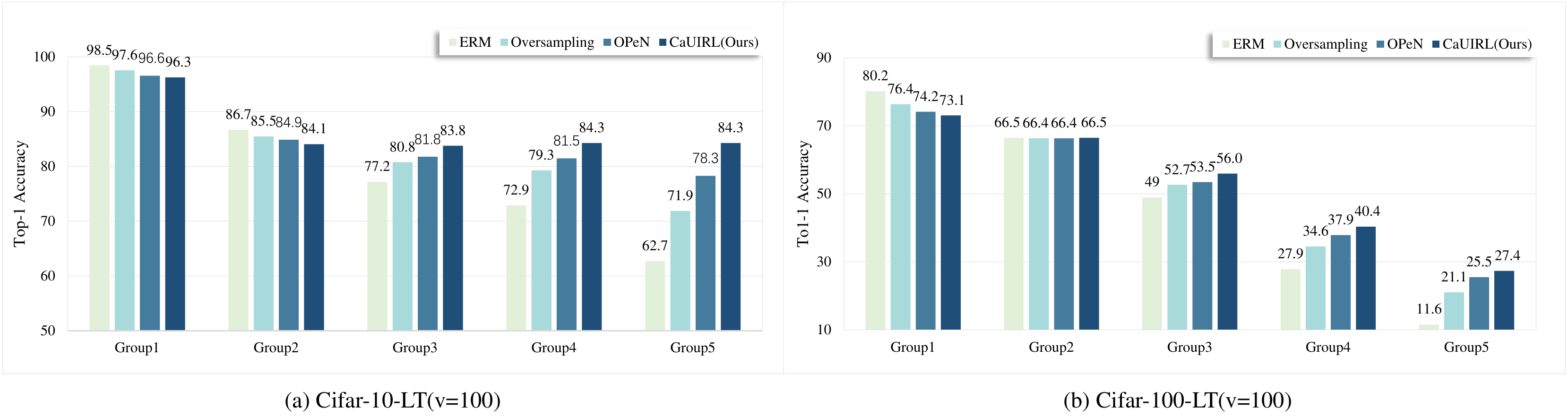}
   \caption{\textbf{Mean accuracy for each group of classes.} (a) Results on Cifar-10-LT ($v=100$), which has two classes in each group. (b) Results on Cifar-100-LT ($v=100$), which has twenty classes in each group.}
   \label{group}
\end{center}
\end{figure*}
\subsection{Re-balance Analysis}
In this subsection, we use the CaUIRL and ERM algorithms to calculate the C2G value on the Cifar-10-LT (v=50/100) dataset, respectively, to explore the variation in the C2G value. The experimental results are shown in Fig.\ref{rebalance}.

In Fig.\ref{rebalance}, the C2G calculated by ERM gradually increases with the increase of the class index, while the C2G value calculated by CaUIRL first increases slightly, and then decreases significantly. This shows that in the tail class (\emph{i.e.}, class $5\sim 9$), the C2G value can be significantly reduced by using the CaUIRL algorithm. From \emph{Definition 1}, we know that the C2G value can measure the degree of domain shift. Therefore, according to the above description, our algorithm can alleviate the problem of domain shift caused by data imbalance, thereby greatly improving the model performance. It also shows that if the chosen Universum is good enough, then our algorithm can achieve comparable performance to balanced scenarios. Combining \emph{Theory 1}, we verify that our method can achieve re-balanced learning both theoretically and experimentally.

\subsection{Model Accuracy for Different Class Sizes}
In this subsection we exemplify that CaUIRL can considerably minimize the generalization error of minority classes, while preserving the accuracy of majority classes. To achieve this , we independently divide Cifar-10-LT($v=100$) and Cifar-100-LT($v=100$) into five non-overlapping groups of similar size based on class size and report the average accuracy for each group of classes.
Specifically, each group contains $20\%$ of the classes, e.g. for Cifar-10-LT, \emph{Group 1} contains the two largest classes in the training set, while \emph{Group 5} contains the two smallest classes. Similarly, Cifar-100-LT has $20$ classes per group.

Fig.\ref{group} shows the classification results of different algorithms according to the above experimental settings, where ERM, Oversamping (deferred oversampling), and OPeN are consistent with \cite{ref10}. It can be intuitively seen that CaUIRL significantly improves the generalization ability of minority classes, which is consistently higher than other algorithms. In particular, in Group4 and Group5 of Cifar-10-LT, the accuracy is improved by $2.8\%$ and $6\%$ compared with OPeN, and the accuracy is improved by $11.4\%$ and $21.6\%$ compared with ERM. In Group4 and Group5 of Cifar-100-LT, the accuracy is $2.5\%$ and $1.9\%$ higher than OPeN, and $12.5\%$ and $15.8\%$ higher than ERM. And when the number of samples in the minority classes are smaller, the relative performance of CaUIRL is better. This shows that CaUIRL can not only improve the overall accuracy, but also has a more obvious promotion effect for small classes. This further reflects that our model can achieve better re-balancing. However, we discovered that the accuracy of Oversampling, OPeN, and CaUIRL in Group1 all reduced marginally. Because, in comparison to with the vanilla ERM, other re-balancing strategies will have a certain degree damage of majority classes in long-tailed recognition.

\subsection{The Impact of  Mixing Coefficient}

\begin{table}[t]
  \centering
  \caption{Effect of mixing coefficient on accuracy}
    \resizebox{\linewidth}{0.5cm}{
    \begin{tabular}{c|ccccccccc}
    \toprule
    \textbf{$\lambda$} & $0.1$   & $0.2$   & $0.3$   & $0.4$   & $0.5$   & $0.6$   & $0.7$   & $0.8$   & $0.9$ \\
    \midrule
   \textbf{ Accuracy }& $87.5$  & $88.37$ & $88.51$ & $88.94$ & \textbf{$89.07$}  & $89.04$ & $88.7$  & $88.71$ & $88.52$ \\
    \bottomrule
    \end{tabular}}%
  \label{mixing}%
\end{table}%
In this subsection, we discuss the effect of the mixing coefficient $\lambda$ on the classification accuracy. We change the value of $\lambda$ from $0.1\sim0.9$ and record the classification top-1 accuracy, the final results are shown in TABLE\ref{mixing}.

TABLE\ref{mixing} reveals an interesting findings where the value of $\lambda$ reflects the trade-off of sample diversity and domain similarity. The entropy of the generated Universum increases as the value of $\lambda$ decreases, increasing the diversity of samples within a class but decreasing the similarity between domains. When the value of $\lambda$ is larger, the entropy of the generated Universum is smaller, which increases the similarity between domains, but reduces the diversity of samples within the class. When we select $lambda=0.5$ as a compromise between the two, we achieve the best classification accuracy of $89.07$.
~\\

\section{Conclusion}
We revisit long-tailed recognition from the perspective of augmenting contradictory data and propose employing CaU to induce re-balance learning.  In particular, we demonstrate that a classifier learned using CaUIRL is consistent with balanced scenarios according to Bayesian theory. Further, we propose HoMu to automatically generate Universum, which additionally takes the domain similarity, class separability and sample diversity into account. Such auto-generated Universum can greatly advance the development of Universum learning. CaUIRL achieves state-of-the-art results on benchmark datasets and can significantly improve the generalization ability of minority classes.
Furthermore, our work provides a novel solution for long-tailed recognition and can also serve as a general data augmentation method.


%

\ifCLASSOPTIONcaptionsoff
  \newpage
\fi

\end{document}